\documentclass[letterpaper]{article} 
\usepackage{aaai20}  
\usepackage{times}  
\usepackage{helvet} 
\usepackage{courier}  
\usepackage[hyphens]{url}  
\usepackage{graphicx} 
\def\UrlFont{\rm}  
\usepackage{graphicx}  
\usepackage{booktabs} 
\begin{document}
%
\title{Supervised Understanding of Word Embeddings}
\author{Halid Ziya Yerebakan, Parmeet Bhatia, Yoshihisa Shinagawa\\
Siemens Medical Solutions USA\\
65 Valley Stream Pkwy, Malvern, PA\\
\{halid.yerebakan, parmeet.bhatia, yoshihisa.shinagawa\}@siemens-healthineers.com
}
\maketitle
\begin{abstract}

Pre-trained word embeddings are widely used for transfer learning in natural language processing. The embeddings are continuous and distributed representations of the words that preserve their similarities in compact Euclidean spaces. However, the dimensions of these spaces do not provide any clear interpretation. In this study, we have obtained supervised projections in the form of the linear keyword-level classifiers on word embeddings.  We have shown that the method creates interpretable projections of original embedding dimensions. Activations of the trained classifier nodes correspond to a subset of the words in the vocabulary. Thus, they behave similarly to the dictionary features while having the merit of continuous value output. Additionally, such dictionaries can be grown iteratively with multiple rounds by adding expert labels on top-scoring words to an initial collection of the keywords. Also, the same classifiers can be applied to aligned word embeddings in other languages to obtain corresponding dictionaries. In our experiments, we have shown that initializing higher-order networks with these classifier weights gives more accurate models for downstream NLP tasks. We further demonstrate the usefulness of supervised dimensions in revealing the polysemous nature of a keyword of interest by projecting it's embedding using learned classifiers in different sub-spaces. 

\end{abstract}

\section{Introduction\label{introduction}}

    Semantic representation of the words is the key ingredient to improve the performance of modern  NLP systems. Mikalov et. al. (\citeyear{mikolov2013distributed}) presented word2vec algorithm to learn word representations from unsupervised corpora via language modeling tasks. Learned word representations preserve co-occurrence relationships in terms of cosine similarity. These dense representations are more compact than prior bag of word methods \cite{joachims1998text},\cite{mccallum1998comparison}  while providing transfer learning capability, thanks to learned similarities across the words. However, since the word vectors are obtained in an unsupervised setting, the dimensions are not well aligned with the final task. Thus, in practice, neural networks are designed by adding layers on top of pre-trained word embeddings. 
\begin{figure}[ht]
\vskip 0.1in
\begin{center}
\centerline{\includegraphics[width=\columnwidth]{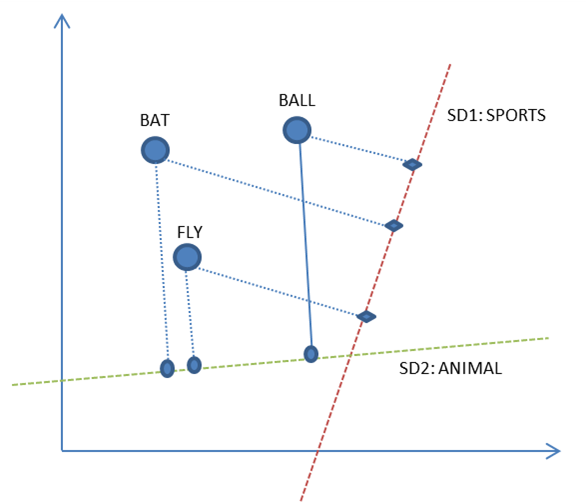}}
\caption{Visualization of our proposed method on toy example. The words lying in two dimensional embedding space are projected into supervised dimensions for topic sports and animal.}
\label{supervised_dimensions}
\end{center}
\vskip -0.2in
\end{figure}    
Although, the representations in the new layers are optimized for the desired task, they are also not easily interpretable. In the case of simple linear dense layers, some dimensions could be observed as near-duplicates in terms of word activations while other dimensions activate on very rare words. This interpretation is more difficult when LSTM or CNN architectures are used. Additionally, transferring them to different tasks is difficult due to their highly optimized nature. In our study, we investigated the semantically meaningful projections in the embedding space. These projections are learned by keyword level supervision by labeling a few keywords.  Thus, we used the term  "supervised dimensions" to refer to these projections throughout the paper. 

Figure \ref{supervised_dimensions} gives an overview of supervised dimensions with the help of a toy example. In this figure, the words originally lie in two-dimensional embedding space. They are projected into two supervised dimensions for sports and animal respectively. These dimensions provide a direct interpretation of target words. In this particular case, SD1 is trained for keywords related to sports and hence the projection of word \textbf{bat} is closer to \textbf{ball} in this dimension compared to the projection of word \textbf{fly}. On the other hand, SD2 is trained for keywords related to the animal category. Hence for this dimension, projections of \textbf{bat} and \textbf{fly} are closer to each other compared to \textbf{bat} and \textbf{ball}. This toy example further demonstrates that supervised dimensions can be used to detect the polysemous nature of the words. In the original embedding space, \textbf{bat} is equally close to the words \textbf{ball} and \textbf{fly}. Whereas, supervised dimension moves the word \textbf{bat} closer to either \textbf{ball} or \textbf{fly} depending on whether it represent topic sports or animal respectively. In other words, specific projections of the embedding space reveal different senses of the same word.

The benefit of having this supervision to obtained projections is keeping dimensions aligned in a semantically coherent way. One such interesting by-product is to obtain groups of semantically similar words which we call "semantic dictionaries". The prediction on remaining word embeddings gives the probability of words belonging to the same class of the words in the positive labeled set. Note that as opposed to topic models the probabilities on words are not normalized  \cite{blei2003latent}, hence each supervised dimension can be treated independently and does not influence the order of words in any other projection. Sorting the words according to probabilities creates an opportunity to refine supervision further to curate a semantic dictionary on the target topic. We also note that the dictionaries, in this case, could be represented with a single vector in embedding space. Owing to its similarity to topic modeling, supervised dimensions can be further used to obtain document representations in this curated semantic representation space. By aligning words embeddings in another language, it is further possible to obtain semantic dictionaries in that language without necessarily retraining supervised dimensions. We have demonstrated this capability by aligning word embeddings of English and German language. The supervised dimensions are learned in English and semantic dictionaries are curated automatically in German language. Due to their coherent structure, supervised dimensions are easily transferable to different tasks. For example, it is straightforward to initialize trainable layers of higher-order networks using the learned weights of these linear projections. In this study, we have experimented addition of supervised dimensions on smoking status identification from radiology reports. We have observed that supervised dimensions improve the F1 score in various network configurations.

	The rest of the paper is organized as follows. In the next section, we briefly review related work and identify their similarities and differences to our method. Section 3 describes the method and implementation details. Later, we demonstrate the applications and insights of supervised dimensions. We finally conclude the paper with some discussions and provide future directions.

\section{Related Work}

    Thanks to very large publicly available corpora, unsupervised learning of word representations capture comprehensive relationship metric across the words. Utilizing these representations on downstream NLP tasks provides additional gains in the evaluation measures \cite{Kim14}. Unsupervised training is usually achieved by language modeling related tasks. In particular, word2vec \cite{mikolov2013distributed} learns word vectors that can predict the central word from the neighboring words within a certain distance from the central word, or that can predict neighboring words from central word using a single layer neural network. In later studies, more advanced models are proposed to produce contextual word embeddings that have been shown to perform better in subsequent tasks \cite{peters2018deep,devlin2018bert,yang2019xlnet}. 
    
    In practice, trainable layers are added on top of pre-trained embeddings. These trainable layers transform word embeddings in highly non-linear fashion for the final task at hand. As a result, added layers of the network activate on a specific portion of the original word embedding space. These activations correspond to sub-set of words in vocabulary that could be treated as topics or dictionaries. However, there is no explicit control over these features. To have the desired control on activation behavior lexicon dictionaries could be curated. Words could be added or removed in this case. However, collecting a dictionary for the required task is a time-consuming process. Thus, semi-automatic ways of curation based on word embeddings have appeared in the literature. For example, the automation of suggesting new keywords into a dictionary named  as term set expansion was suggested by \cite{mamou-etal-2018-setexpander}. In our setting, we are using a simple linear logistic classifier to train each feature. In this study, we investigate different properties of these projections and the utilization of the projections in a higher-order network. 
    
    Our method is also related to topic models. Similar to topic models, supervised dimensions provide features to represent documents at a more abstract level \cite{blei2003latent}. To obtain topics, topic models are trained in unsupervised fashion on a text corpus. During inference, computed topics can be used to represent documents in the test set. However, these topics are generally not optimized for the final task. For example, medical words may be grouped into one topic, but cardiovascular diseases will not likely have an independent topic. In other words, it is very difficult to control the model to have a dedicated topic for cardiovascular diseases. Increasing the granularity of the topics not only reduces the quality of desired topics but also introduces many unwanted topics as well. In a recent study Dieng et. al. \shortcite{dieng2019topic} demonstrated a generative topic model on word embedding spaces . They have utilized the log-linear model with topics in word embedding space. However, it inherits the unsupervised nature of LDA topic models. 
    
    Post-processing of word embeddings for various purposes is considered in the literature. Unsupervised dimensionality reduction methods via autoencoder or PCA, \cite{raunak-etal-2019-effective,Tissier2018NearLosslessBO} approaches have been utilized. In our study, we focus more on utilizing supervision information to extract interpretable dimensions. 
    

%
%
%
%
%
%
%
\section{Iterative Supervision}

\begin{figure}[ht]
\vskip 0.1in
\begin{center}
\centerline{\includegraphics[width=\columnwidth]{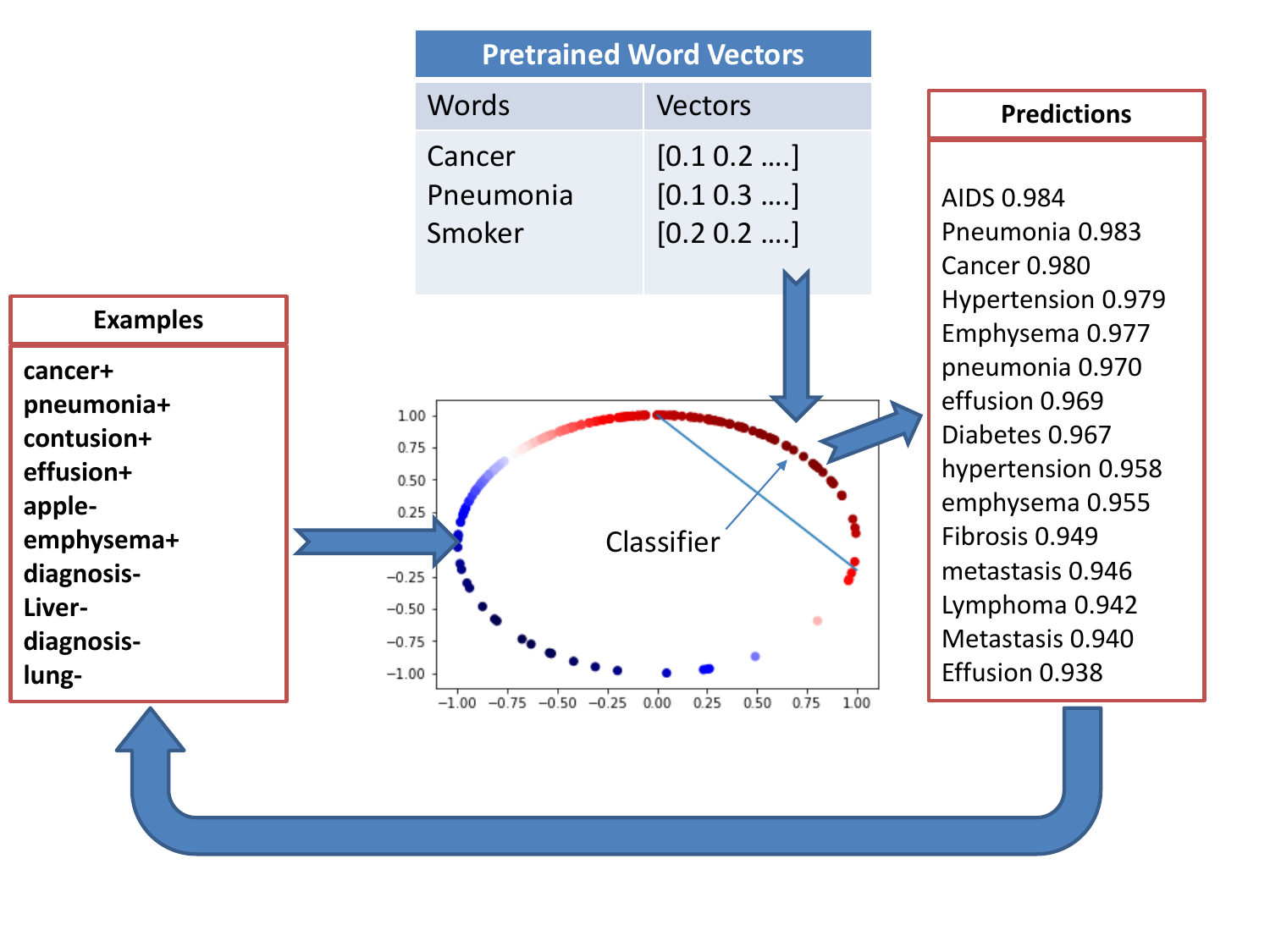}}
\caption{Active learning method to learn supervised linear projections of embedding space}
\label{online_supervision}
\end{center}
\vskip -0.2in
\end{figure}


In our setting, we propose to use linear projections of word embeddings by training binary classifiers. These classifiers are trained with active learning supervision on normalized vectors as shown in Figure \ref{online_supervision}. The normalization ensures that the linear classifier produces similar results on the words having higher cosine similarities. Therefore, words having similar themes to positive words will have higher scores according to the classifier. Also, since the normalized word vectors lie on a hyper-sphere, each classifier corresponds to a hyper-cap geometrically. Logistic regression is used as the classifier model. However, any other linear model would be equally applicable.

		In the initial stage of the algorithm, a few positive keywords that belong to the same topic are provided. In addition to the positive samples, randomly sampled negative words are added to train a binary logistic regression. Then, the logistic regression classifier provides the prediction probabilities on each of the remaining words in the vocabulary. After sorting these words, the top-scoring words are added to the initial set of positive keywords. Subsequent iterations usually provide negative examples as well. This process is repeated for a few iterations to finally obtain the classifiers. Both generalization level of the projection and theme of the projection is managed by adding positive or negative keywords. 
		
		

%
%

\section{Applications of Supervised Dimensions}
In this section, we have demonstrated different applications of supervised dimensions. In our experiments, we have used scikit-learn linear logistic regression model with a positive class weight of 2 to enhance the effect of positive words. We have used top 250k words of Fasttext Common Crawl word embeddings \cite{fasttext} after L2 normalization. 

\subsection{Transfer Learning\label{network}}
\label{transfer_learning}

\begin{figure}[!htbp]

\vskip 0.1in
\begin{center}
\centerline{\includegraphics[width=0.8\columnwidth]{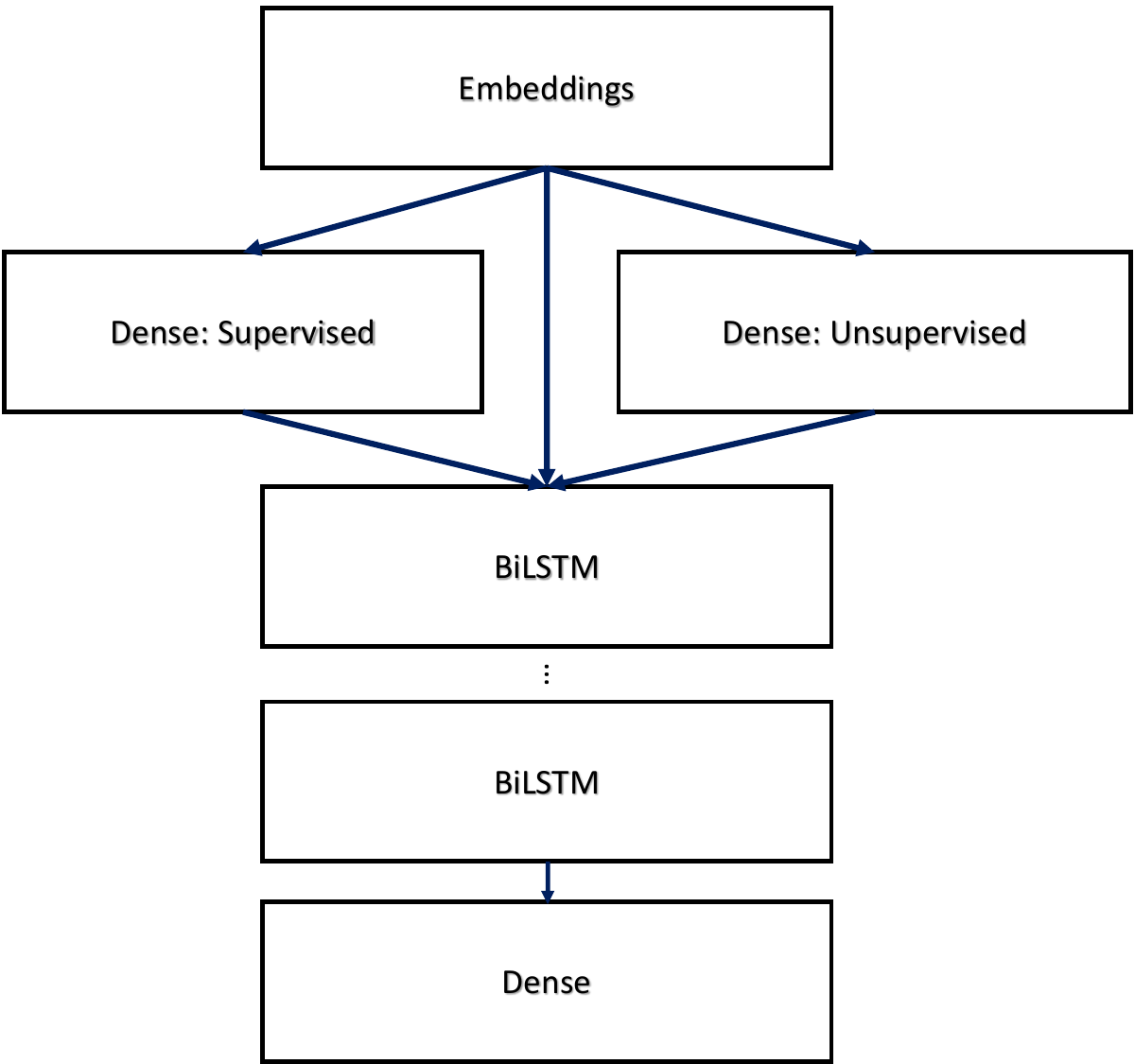}}
\caption{Multi-layer BiLSTM network with supervised and unsupervised dimensions\label{bilstm}}
\end{center}
\vskip -0.2in
\end{figure}

Transfer learning is a common practice in standard NLP tasks where-in the weights of some or whole of the neural network are learned on tasks for wherein large datasets are readily available, for instance, language modeling.  Subsequently, the network structure is modified to meet the needs of downstream tasks and they are fine-tuned with task-specific loss functions. In our network structure, we have used multi-layer BiLSTM on top of concatenated transformed embeddings from unsupervised and supervised dense layers as shown in Figure \ref{bilstm}.  Trained logistic regression functions could be used as nodes in this network by replacing the weights of dense layers with that of learned keyword classifiers. Activation functions could be changed since the underlying linear initialization provides a necessary base for subsequent training. We observed that allowing these weights to be trainable provides additional performance gains. Besides the supervised dimensions, adding randomly initialized unsupervised dense layer improves the overall accuracy.  Dropout is utilized to regularize the network.

We have demonstrated the effectiveness of the proposed scheme on smoking status identification task in medical report sentences. In this problem, we classified sentences that are relevant to smoking into four different smoking status namely "previous smoker", "non-smoker", "current-smoker" and "unknown". Relevant sentences are selected by certain keywords that are indicative of smoking status from the reports. NLTK word tokenizer is used for the tokenization of the report. We measured F1 score to assess the performance for standard BiLSTM network, BiLSTM network with unsupervised dense layer on top of word embeddings and BiLSTM with a combination of supervised and unsupervised dense layers.  Our dataset contains 3846 examples. We used 5-fold cross-validation with 20 repetitions and reported the average for each configuration. Adam optimizer with a learning rate of $10^{-3}$ is used as an optimization method. 

\begin{figure}
\vskip 0.1in
\begin{center}
\centerline{\includegraphics[width=\columnwidth]{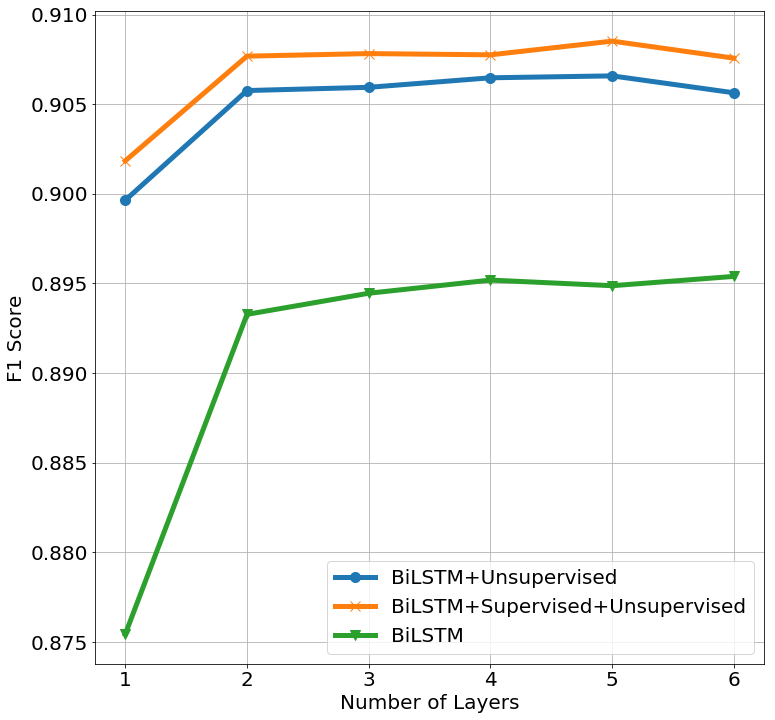}}
\caption{Change in classification F1 under different network configurations}
\label{valid_accuracy}
\end{center}
\vskip -0.2in
\end{figure}

We have increased the number of BiLSTM layers and have measured the performance of three architectures across all the settings. As can be seen in Figure \ref{valid_accuracy}, just adding more BiLSTM layers does not help to improve the accuracy notably after the second layer.  On the other hand, adding a dense layer on top of word embeddings, provide a boost on performance. Furthermore, initializing dense layers with weights of the linear keyword-level classifiers provides additional performance gains suggesting the transfer learning capability of supervised dimensions. 

The dense layer activations in this experiment can be precomputed since every word corresponds to a unique value. Thus our method can be considered as augmentation of word embedding dimensions via informative interpretable projections.

\subsection{Dictionary Curation}

    One can directly benefit from proposed supervised dimensions by allowing curation of a topic-specific dictionary. This can be achieved by getting predictions of the words in the embedding database as described in the previous section. Later, a threshold value can be used to get a complete word list.  These vocabularies can be used in keyword searches or named entity recognition tasks later.

    The central question for dictionary curation is how accurately the learned linear classifiers can predict the unlabeled part. To conduct a performance evaluation on these classifiers, we kept 10\% of the annotated keywords for validation. We have calculated F1 scores for 30 different classifiers based on our data and calculated micro and macro averages as shown in Table \ref{performance}. The results suggest that our proposed method can be used to create topic-specific dictionaries with highly coherent structure among words. 

\begin{table}[!htbp]
\caption{Data statistics and performance of the dictionary curation}
\label{performance}
\vskip 0.1in
\begin{center}
\begin{small}
\begin{sc}
\begin{tabular}{lc}
	\toprule
	\# Positive Examples    & 32686  \\
	\# Negative Examples & 15774 \\
	Micro Average F1     & 0.9102 \\
	Macro Average F1      & 0.8775\\
	\bottomrule
\end{tabular}
\end{sc}
\end{small}
\end{center}
\vskip -0.1in
\end{table}

\subsubsection{Dictionary Curation on Multiple Languages}

\begin{table*}
\caption{Multi language dictionary curation}
\label{multilanguage}
\vskip 0.1in
\begin{center}
\begin{small}
\begin{sc}
\begin{tabular}{ll}
	\toprule
	EN & smoking, smoker, tobacco, smokers, smoke, cigarettes, cigarette, Smoking \\
	DE & Rauchen, rauchen, Raucher, geraucht, Rauchens, Zigaretten, Zigarette, raucht \\
	NL & roken, roker, sigaretten, gerookt, rookt, sigaret, tabak, Roken \\
	\bottomrule
\end{tabular}
\end{sc}
\end{small}
\end{center}
\vskip -0.1in
\end{table*}

    Another by-product of supervised dimensions is getting dictionaries in other languages without retraining the classifiers. The trained classifier in one language works on multiple languages if the underlying word embeddings are aligned. To demonstrate this, we have trained a classifier with three positive words "smoking", "smoker" and "tobacco" in English and we have used the same classifier on aligned word embeddings of German and Dutch. We have sorted words based on their probabilities. The top 8 candidates are shown in Table \ref{multilanguage}. All words displayed in the table are relevant to smoking. It can be seen that supervised dimensions could reduce the efforts to scale models to multiple languages considerably. 

\subsection{Polysemy Representations}

\begin{table*}
\caption{Projection with trained classifiers uncovers hidden polysemy in word vectors\label{pol}}
\vskip 0.1in
\begin{center}
\begin{small}
\begin{sc}
\begin{tabular}{ll}
	\toprule
	play + bat + run   & playing, runs, plays, played, running, paly, game, go  \\
	play + script + art  & scripts, scripting, plays, playwriting, script., theatre, playing, artwork \\
	bat + animal + fly & bats, flies, bird, flying, mammal, animals, insect, Bat\\
	\bottomrule
\end{tabular}
\end{sc}
\end{small}
\end{center}
\vskip -0.1in
\end{table*}

    Different projections of the same keyword contain different meanings in the word vector space. In other words, cosine similarities among words are different in sub-spaces.  This can be verified by training specific supervised dimensions with a few keywords. In our experiment, we used 3 positive keywords with random negative samples on 3 supervised dimensions. Top 8 positive candidates after applying projection on the remaining words in the vocabulary are shown in Table \ref{pol}. On the left side, seed keywords are shown. For example, the word "play" means a baseball-related concept in the first line whereas it corresponds to art in the second. Both dimensions will be activated for the occurrence of the word "play" in the input text indicating its polysemous character.  Similarly, the word "bat" provides different top scoring words in the context of "animal" and "play" respectively again indicating its polysemous nature.
    


\section{Interpretation of Supervised Dimensions}
In this section, we will demonstrate the interpretability of supervised dimensions with the help of a few concrete examples.

\subsection{Activated words in Dense layer outputs}

\begin{table*}[!htbp]
\caption{ Supervised dimensions keep the topics consistent while unsupervised dimensions are not easily interpretable }
\label{inter}
\vskip 0.1in
\begin{center}
\begin{small}
\begin{sc}
\begin{tabular}{ll}
	\toprule
	Current   & continues, constant, constantly, continuous, frequent, continuously, continual, ongoing \\
	Family  & mother, father, grandmother, dad, Grandmother, grandfather, mothers, grandchild\\
	Negation & anyting, cant, wont, no, Cannot, cannot, nothing, anthing, doesnt, wouldnt \\
	Unsupervised  & whiner, meanie, youngling, Fro, cranky, Kalos, tummy, Whiny  \\
	Unsupervised   & readout, FS, TRK, HV, LV, Transponder, Heading, Medium, FLX, Go-to  \\
	Unsupervised   & help, Help, Molo, stabilize, expanding, envision, trillion, Brainstorm \\
	\bottomrule
\end{tabular}
\end{sc}
\end{small}
\end{center}
\vskip -0.1in
\end{table*}

To check the coherency of supervised dimensions, we investigated dense layer outputs and collected words that trigger various dimensions in dense layers. For this task, we have used the same network trained for classification task in the smoking status identification experiment. We selected three relevant dimensions from the dense layer initialized with learned classifier weights and identified the top scoring words for those dimensions. Similarly, we identified the top scoring words from three output nodes of the dense layer with random initialization (unsupervised). 

\begin{figure}[ht]
\vskip 0.1in
\begin{center}
\centerline{\includegraphics[width=\columnwidth]{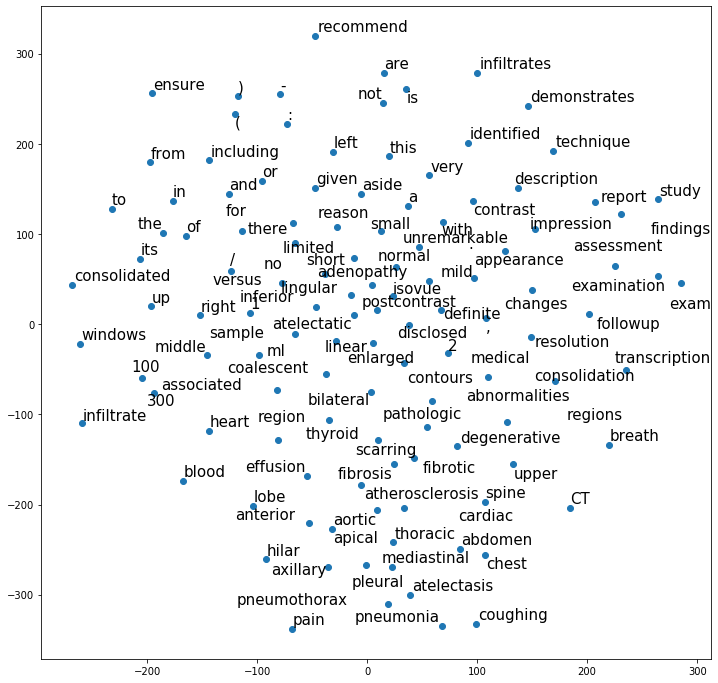}}
\caption{t-SNE plot using original word embeddings}
\label{before_training_tsne}
\end{center}
\vskip -0.2in
\end{figure}

We observed that supervised dimensions provide more coherent topics as shown in first three lines of the Table \ref{inter}. Interestingly, even after fine-tuning with downstream classification task, those dimensions mostly preserve the original topic for which they were trained. On the other hand, unsupervised dimensions provide no coherence structure among words activated.

\subsection{t-SNE Visualizations}

\begin{figure}[ht]
\vskip 0.1in
\begin{center}
\centerline{\includegraphics[width=\columnwidth]{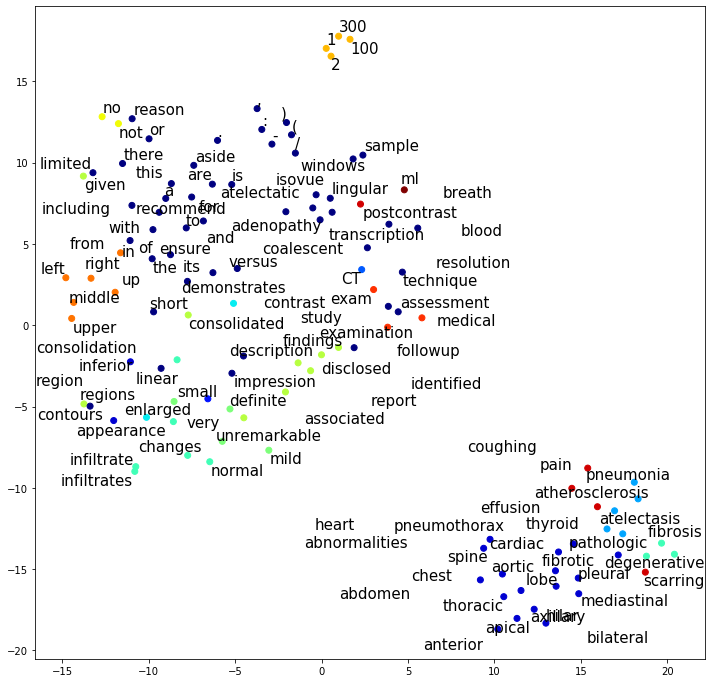}}
\caption{t-SNE plot using supervised dimensions}
\label{after_training_tsne}
\end{center}
\vskip -0.2in
\end{figure}

Another interesting direction to understand interpretability of supervised projections is using t-SNE plot. We have chosen an example chest report from mtsamples \footnote{\url{https://www.mtsamples.com/}} and taken the unique words in the report\footnote{Stop word removal is not necessary since each word is taken only once in this experiment.}. The Figure \ref{before_training_tsne} shows the t-SNE visualization of word embeddings in the original embedding space with perplexity value of 15.  The embeddings are preserving similarities, i.e similar words are close to each other in low dimensional projection using t-SNE. However there is no significant cluster that emerges in the projections. On the other side, as show in Figure \ref{after_training_tsne} for t-SNE projections of supervised dimensions, the words are nicely grouped according to classifiers which are represented in different colors. The words are assigned to the classifier that correspond to the maximum value of activation. Some related groups are clustered together e.g. symptoms, diseases and anatomies. For some words in the report, the maximum activation value is below .5. They are not assigned to any classifier and are shown with dark blue color in the plot. We further note that they comprise of stop words mostly for which we did not train any explicit classifier.

\subsection{Topic modeling}
\begin{figure}[ht]
\vskip 0.1in
\begin{center}
\centerline{\includegraphics[width=\columnwidth]{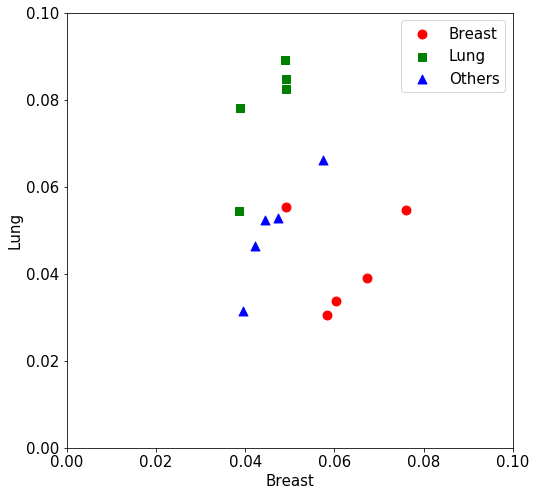}}
\caption{Scatter plot of medical reports using two supervised dimensions}
\label{topics}
\end{center}
\vskip -0.2in
\end{figure} 

As we have shown in the previous subsection, we can obtain semantic representation for every word by projecting embeddings into supervised dimensions in a document. Similarly, we can compute the document-level semantic representation by taking the average representation of the words in that document. To demonstrate this method, we investigated the representations of medical corpora using two supervised dimensions. We have sampled 5 lung radiology reports, 5 breast radiology reports, and 5 other medical reports from mtsamples. Then, we have trained the supervised dimensions for the breast and lung-related topics with 25 positive keywords. We calculated the logistic regression classification probabilities for every word in the reports while removing stop words. Lastly, we have taken the average of these probabilities in each report. As a result, every report is represented by two dimensions. 

	The resulting representations are shown in Figure \ref{topics}. The transverse axis of the plot corresponds to the breast feature value, and the vertical axis is the lung feature value. As can be seen in the figure, the reports can be mostly clustered in the three groups corresponding to breast, lung, and the other. To visualize the ground truth labels, we colored the points corresponding to the lung reports in green, breast in red and the others in blue. The lung feature is discriminative for the lung reports. The breast feature is generally more active, only failing in the first report. When we looked at this first report, there are mentions of the chest which typically appear in lung reports as well. Similarly, the "other" reports are mostly inactive for these two features.





\section{Conclusion}

	We have investigated the use of linear supervised keyword classifiers on word embeddings to obtain meaningful projections of word embeddings in a low dimensional space. We have shown that these supervised dimensions are not only more interpretable but can also be used in transfer learning to get additional gains in downstream NLP tasks. Furthermore, as these classifiers can be trained iteratively in an active learning framework, it can be used to create lexicon dictionaries. Finally, we demonstrated interpretations of the supervised dimensions on word activations and document representations.
	
One of the future directions is to investigate the use of supervised dimensions to reduce the size of the embedding matrices for model compression. Also, our analysis shows that explicitly contextualized word embeddings are not necessary to encode polysemy. However, contextualized word embeddings can create conditional activations that could help to detect phrases. 

We would further like to inform our readers that the presented work is for research, and neither for diagnostic use nor use in products.
\bibliography{superviseddimensions}
\bibliographystyle{aaai}
\end{document}